\documentclass{article}
\usepackage{ijcai22}

\usepackage{times}

\usepackage{soul}
\usepackage{url}
\usepackage[hidelinks]{hyperref}
\usepackage[utf8]{inputenc}
\usepackage[small]{caption}
\usepackage{graphicx}
\usepackage{amsmath}
\usepackage{booktabs}
\usepackage{xcolor}
\usepackage{wrapfig}

\everypar{\clubpenalty=0}
\everypar{\widowpenalty=0}
\everypar{\displaywidowpenalty=0}
\everypar{\looseness=-1}
\newcommand{\rita}[1]{\textcolor{black}{#1}}

\title{ $<$ $<$ Beware the Rationalization Trap!  $>$ $>$\\ 
\textit{\large When Language Model Explainability Diverges from our Mental Models of Language} }

\author{
Rita Sevastjanova$^1$\and
Mennatallah El-Assady$^2$
\affiliations
 $^1$University of Konstanz\\
 $^2$ETH Zürich\\
\emails
 rita.sevastjanova@uni-konstanz.de,
 melassady@ai.ethz.ch
}

\begin{document}

\maketitle

\begin{abstract}
Language models learn and represent language differently than humans; they learn the \textit{form} and not the \textit{meaning}.
%
%
Thus, to assess the success of language model explainability, we need to consider the impact of its divergence from a user's mental model of language. 
In this position paper, we argue that in order to avoid harmful rationalization and achieve truthful understanding of language models, explanation processes must satisfy three main conditions: (1)~explanations have to truthfully represent the model behavior, i.e., have a high fidelity; (2)~explanations must be complete, as missing information distorts the truth; and (3)~explanations have to take the user's mental model into account, progressively verifying a person's knowledge and adapting their understanding. 
We introduce a decision tree model to showcase potential reasons why current  explanations fail to reach their objectives. 
We further emphasize the need for human-centered design to explain the model from multiple perspectives, progressively adapting explanations to changing user expectations.  
\end{abstract}

\section{Introduction}
Machine learning models are widely used for diverse applications and use cases, such as machine translation, image recognition, patient diagnostics, etc. 
These models are typically black boxes as they do not provide any information considering how they evolve nor how they make their predictions.
Thus, the Explainable Artificial Intelligence (XAI) research field~\cite{arrieta2020explainable} 
has grown rapidly in recent years. 

One specific area of interest is modeling and understanding natural language. A wide range of deep-learning-based language models (LMs) (e.g., BERT~\cite{devlin2018bert}) have been developed in recent years, reaching high performances in various natural language understanding and generation tasks. 
These models are pre-trained on a large number of text documents (e.g., \textit{Wikipedia} articles), and are capable of learning various language structures in an unsupervised manner. 
The popularity of deep-learning-based language models and their widespread usage in diverse natural language processing (NLP) applications for prediction and decision-making~\cite{vig2019analyzing} drove the necessity for their explainability.
In this paper, we want to emphasize the \textbf{challenges related to language model explainability}. 
%


Language (text data) is \textbf{\textit{deceivably} intelligible}. We all have an inherent understanding of language, as this is one of the most effective communication mediums for humans. Most people are often under the illusion they understand all aspects of language. However, the reality is that we usually understand the text \textit{semantics} (intelligible) but not as much the text \textit{structure} and linguistic \textit{function} (non-intelligible). Unless we specifically analyze text data,  we are usually interpreting language based on our intuition and experience. Similarly to other habitual and intuition-based tasks (think: car driving, for example), understanding language is something that humans do not  consciously or effortfully \textit{think} about.  
However, this is in stark contrast to what we need for understanding language models. These models process text data in a sequential and logical manner, scouting patterns to learn decision boundaries. These often enough do not align with human intuition, making them ideal candidates for \textit{harmful rationalizations}.

Hence, the first major challenge in explaining language models is the misalignment between how language models learn and represent language and the way humans perceive it (through their intuition). In particular, language models currently learn only parts of the semantic and functionality spectrum~\cite{dasgupta2018}. 
Thus, also explanation methods might not be faithful to the model's decision-making and are likely to be incomplete. 
Humans typically tend to ``fill the missing gaps'' of the explanations by relying on their world understanding. This is dangerous, since such a rationalization can lead to wrong conclusions about the model's behavior.
Therefore, it is especially important to think about the \textbf{quality} and \textbf{completeness} when designing new explanation methods. 

The second major challenge is effectively tailoring explanations to specific stakeholders.  
It has been shown that current language model explanation methods have limited success when it comes to increasing user understanding of the model's behavior~\cite{arora2021explain,Hase2020EvaluatingEA}. 
Since there are several user groups involved in the explainability process (e.g., NLP researchers with theoretical or computational background, decision-makers, etc.), they require different types of explanations matching their expectations.
Each group of users has its specific mental model of language, based on their background, education, or world experience. 
To reach a positive explainability outcome, it is thus important to design human-centered explanations that align with the different \textbf{human mental models of language}.

To summarize, in this position paper we argue that addressing user mental models (i.e., their understanding and use of language) is especially important for the NLP explainability, since humans have a strong mental model of language based on their experience.
And when we design or consume explanations, we sometimes forget that language models learn only the form and not meaning~\cite{bender-koller-2020-climbing}. 
In this paper, we show that the success of explanations, i.e., \textit{truthful understanding of model behavior}, depends on three main conditions -- \textbf{explanation quality},  \textbf{completeness}, and their \textbf{fit to user's mental models}.
Only with complete, high fidelity explanations that match or manage to calibrate user mental models can we guarantee to not fall into a \textit{harmful rationalization trap}. 

\section{Background: Human-Centered XAI}

In the following, we describe the types of current explanation methods and the human role in the explanation process.

\subsection{Types of Explanations}
Most commonly, explanations in XAI are categorized, as follows  -- (1) whether the explanation is for an individual prediction (local) or the model’s prediction process as a whole (global), and (2) whether the explanation is emerging directly from the prediction process (self-explaining) or it requires post-processing (post-hoc)~\cite{guidotti2018survey}. Additionally, we can distinguish between (3) explanations on the data-level (model-agnostic) or explanation taking the model behavior into account (model-aware)~\cite{spinner2020explainer}. Lastly, (4) explanations can be top-down (deductive), bottom-up (inductive), or contrastive~\cite{strategies2019}. 
Regardless of their type, the objectives of these explanations are clear -- they should faithfully characterize the models’ behavior (i.e., be faithful), and increase user understanding and trust in black-box models~\cite{jacovi-goldberg-2020-towards}. They should be interpretable, accurate, and with a high fidelity, i.e., the model should be able to accurately imitate a black-box predictor~\cite{guidotti2018survey}.

\subsection{Human Role in the Explanation Process}
The users to whom the explanations are provided play a crucial role in the explanation process and in its success. 
Yet, it has been criticized that this role is often underestimated.
\cite{ehsan2020human} recently wrote that ``\emph{explainability in AI is as much of a Human-Computer Interaction (HCI) problem as it is an AI problem, if not more. Yet, the human side of the equation is often lost in the technical discourse of XAI.}'' 
Also,~\cite{liao2021human} emphasize that the choice of an explanation method should be based on target users’ explainability needs, whereby ``\emph{their explainability needs can vary significantly depending on their goals, backgrounds, usage contexts, and more.}''
So, who are the users for whom the explanations are designed?
In the literature, one can find five main user groups for model explainability: (1) \textbf{model developers}, (2) \textbf{business owners or administrators}, (3) \textbf{decision-makers}, (4) \textbf{impacted groups}, (5) \textbf{regulatory bodies}~\cite{arrieta2020explainable}.
Based on their needs, each group might prefer a different type of explanation.
At the same time, there exist from user groups independent, fundamental
properties of explanations. ~\cite{miller2019explanation} in his survey summarized their four major properties: (1) \textbf{explanations are contrastive}, i.e., people do not ask why an event happened, but rather why \textit{it} happened instead of \textit{something else}; (2) \textbf{explanations are selected} by humans, and they rarely expect explanations to consist of a complete cause of an event; (3) \textbf{explanations are social}, and they are part of a conversation; and (4) \textbf{probabilities matter less} than causal relationships. 

One of the objectives of good explanations is their faithful characterization of a models’ behavior. 
When these explanations are fragmented, i.e., they partially explain what the model learns or when they don't fully match one's expectations, people start to rationalize about the explanations as well as the explained model. 
\textit{Rationalization} is an attempt to find reasons for behavior, decisions, etc.~\cite{vanhoucke_2018}
If explanations miss relevant aspects for a complete understanding, users can make false conclusions about the model's behavior and explainability process fails.

\section{Language Model Explainability}

With the emergence of Transformer architecture~\cite{vaswani2017attention}, there was a paradigm shift away from building NLP applications on, e.g., Recurrent-Neural-Networks (RNNs)~\cite{bengio1994learning} or Long Short-Term Memory networks (LSTMs)~\cite{hochreiter1997long} and the focus was put on (self-) attention mechanisms.
The self-attention mechanism processes an input word, accounting for its context, in a bidirectional manner~~\cite{vaswani2017attention}. 
Similar to other deep learning models, language models are often fine-tuned with additional output layers for classification tasks (see~\cite{zhuang2020comprehensive}). 
The main idea is to first pre-train the model on large-scale corpora and then fine-tune it on an additional, task-specific labeled corpus~\cite{howard-ruder-2018-universal}. 
And, hence, these models are widely applied in different domains for prediction- and decision-making (e.g., \cite{Sun2019HowTF,Han2019UnsupervisedDA}).
Researchers in the explainable NLP field actively analyze which characteristics these models capture both during the pre-training, and after they are fine-tuned for specific tasks (see, e.g., \cite{rogers2020primer,Danilevsky}).

\rita{Language models are trained on plain natural language; they have no additional input about word classes, their semantic meaning, or functionality.
During the training process, the model represents a word through a contextualized word embedding -- word vector that is sensitive to the context in which it appears~\cite{naseem2021comprehensive}. 
\cite{ethayarajh2019contextual} and \cite{sevastjanova2021} have explored the embedding contextualization, i.e., how they change within the model's architecture, 
and showed that the contextualization increases with language model's layers. 
It has also been shown that in BERT's upper layers, words within the context (i.e., sentence) become more similar to each other~\cite{ethayarajh2019contextual}.
}

\noindent\textbf{Limitations of Current Explanation Methods --} 
In a recent survey, \cite{human-centered-exnlp:2021:hxcai} present an overview of methods for NLP explainability, summarizing and tagging over 200 papers that cover related topics. 
Their work shows that most popular explanation methods in NLP are \textit{feature attribution} (sometimes called \textit{saliency} or \textit{importance}) methods, followed by \textit{tuple}, \textit{rule}, or \textit{concept} formats to demonstrate the model’s reasoning process. 
Less than 1\% use \textit{counterfactuals} -- methods that according to the theory are preferred by humans~\cite{miller2019explanation}. 
However, the feature attribution methods that are used in around 44\% of related papers, are statistical explanations and, theoretically, less sufficient for humans than, for instance, causal explanations~\cite{miller2019explanation}. 

The popular feature attribution methods are local post-hoc explanations that show the importance of a token with respect to the model's prediction. 
A higher score denotes greater importance. 
Example methods include integrated gradients~\cite{sundararajan2017axiomatic} and LIME~\cite{lime}.  
These scores are often visualized in saliency maps in the form of heatmaps (e.g.,~\cite{tenneyGoogleLIT,sinha2021perturbing}).
It is assumed that users can easily understand why models make their decisions~\cite{sun2021interpreting} or when models pay attention to the wrong words~\cite{fleisher2021understanding} by examining the most salient part(s) in the text.

According to the interpretable NLP systems objectives~\cite{jacovi-goldberg-2020-towards}, saliency visualizations should faithfully characterize the important features and increase the user's understanding of the language model's behavior. 
However, current studies show that saliency visualizations \textbf{do not} help users to understand or simulate the model or flip its output~\cite{arora2021explain,Hase2020EvaluatingEA}, but when they help -- the improvement in human performance is only minor~\cite{10.1145/3287560.3287590}. 
Although they are commonly used as explanation methods, they do not satisfy the interpretable NLP system objectives.
What could be the reason?

\begin{figure}[b]
    \vspace{-10pt}
    \centering
    \includegraphics[width=\linewidth]{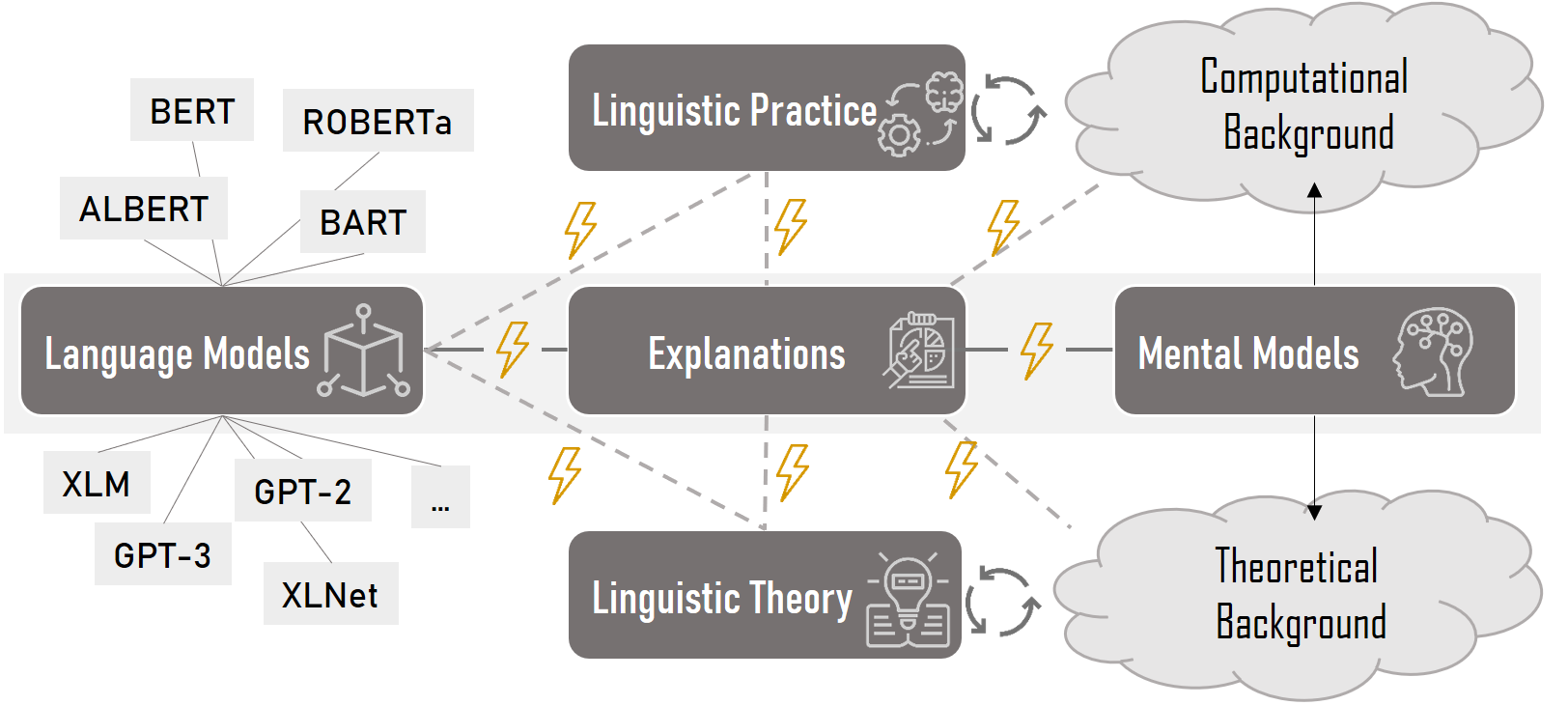}
    \caption{Explainability process in NLP involves three main aspects: language models, explanations, and user mental models. Mental models of language, however, depend on user backgrounds and could be influenced by, e.g., linguistic practice or theory. Since language models learn language differently from humans, there can be many mismatches in the explanation process.}
    \label{fig:model}
\end{figure}

\section[A Thought Experiment  on Explainability]{A Thought Experiment on Explainability}

The goal of explainability is to make a model's behavior understandable to different types of users through XAI methods.
This explainability process in NLP involves three main aspects, as depicted in \autoref{fig:model}. First, we have the language models themselves  with their architectures, behavior, and learning techniques.
Second, we have the various types of explanations by the XAI methods. Third, we have the users and their mental models of language, described in detail in \autoref{sec:menalmodels}. 
Although it would be wishful that these three aspects operate in sync, i.e., the explanations correctly depict model behavior, and the models represent language the way people use it, in reality these aspects are often disconnected. 

The two main issues disrupting this process are, on the one hand, that the fidelity of some explanation methods is still disputed; on the other hand, the lack of human-like concept understanding in language models, causing their learning and results to be widely criticized.
For instance, \cite{bender-koller-2020-climbing} have criticized the hype of language models and scrutinized exaggerated statements, such as, their ability  to \textit{understand} or \textit{comprehend} natural language. 
They argue that a language model that is trained purely on form will not learn meaning -- it simply lacks the signal for it.

However, when people learn, they learn from the world around them and from the interaction with other people in that world~\cite{tomasello1986joint},
i.e., language learning for humans does not happen in isolation, nor based on a single signal.
Since this topic is discussed in detail in context of language model learning (please, see ~\cite{bender-koller-2020-climbing}), we will not go in more detail here.
But, when designing explanation, we need to consider that language models and humans learn language in vastly different ways.

\begin{figure*}
    \centering
    \includegraphics[width=.9\textwidth]{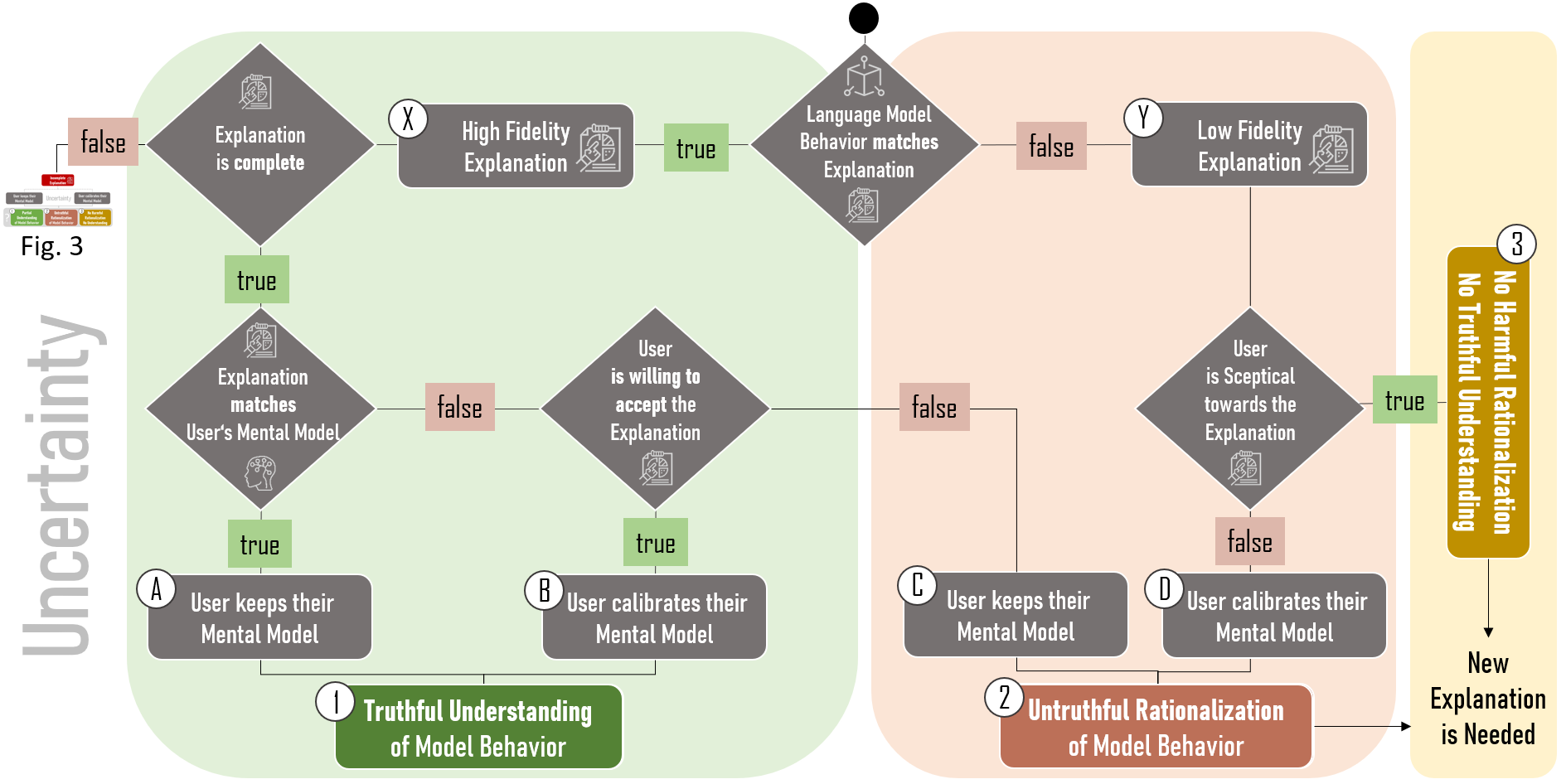}
    \caption{Decision tree showing the explainability process in NLP. (1)~\textbf{Truthful Understanding} of model behavior  can be reached when high fidelity (X), complete explanations match (A) or adapt (B) user mental models. (2)~\textbf{Untruthful Rationalization} begins when explanations have low fidelity (Y) that change user mental models (D), or users are not willing to change their mental models in favor of high-fidelity explanations (C). Only when low-fidelity explanations raise skepticism in users, (3)~\textbf{No Truthful Understanding}
    will be generated.}
    \label{fig:decision-tree}
\end{figure*}

\subsection{Mental Models in NLP}
\label{sec:menalmodels}

Human mental models of language are critical factors influencing the XAI process outcome.\\

\noindent\textbf{Mental Models are Human and Subjective  --} 
According to~\cite{jones2011mental}, ``\emph{Mental models are personal, internal representations of external reality that people use to interact with the world around them.}''
Mental models are crucial in our lives, we use them to reason and make decisions. 
They also provide the mechanism through which we filter and store new information.
Everyone who uses language has a mental model for it. Mental models are subjective, and they depend on a person's background, unique life experiences, perceptions, and understandings of the world~\cite{jones2011mental}, as well as different circumstances in which the models are used. For example, a person's mental model of a car depends on their association to cars in different roles. Being a car designer or mechanic conjures into the mind a complete different conceptualization than being a driver or passenger. \\

\noindent\textbf{People Adapt Their Mental Models --}
In the scope of HCI research field, it has been discussed that mental models evolve as users interact with a system~\cite{norman2014some}.
It is similar with our mental model of language.
During our childhood or education phase, we learn things such as  word semantic meaning, syntactic structures, word functionality in a sentence, and with each new learned concept we update our mental model of language. 
If we acquire  a second language as adults, we extend our  mental model of language, but instead of starting with a blank canvas, we learn in a contrastive manner to our existing understanding. 

\noindent\textbf{Types of Mental Models in NLP --} 
As mentioned earlier, mental models are subjective, and they depend on a person’s background and tasks at hand. 
Thus, also in NLP as a field, researchers and practitioners, depending on their educational backgrounds and experiences, can have different mental models of language.
On the broadest level, we can differentiate between computational and theoretical linguistic backgrounds.
As depicted in \autoref{fig:model}, mental models of researchers and practitioners with computational background are impacted by their linguistic practice, e.g., which algorithms, techniques they use to (pre-) process text data in their research. 
In contrast, mental models of researchers and practitioners with theoretical background are framed according to the linguistic theory they study. 
Although mental models from these two groups of people share a common core, there might also be vast differences between them, for example when judging the importance of specific word categories (e.g., stopwords) for a text analysis task.

\subsection[The Process of Explainability  and Mental Model Adaptation]{The Process of Explainability and Mental Model Adaptation}

As mentioned earlier, the explainability process in NLP consists of three aspects: the language model, the explanations, and the user's mental model. 
The outcome of the explanation process depends on how these aspects align, i.e., whether the explanations correctly explain the model's behavior (high fidelity), whether they match the user's mental models, and if not, whether the user is willing to adapt their mental model in favor of the explanations.

In the following, we describe the explainability process with its components in a form of a decision tree (shown in \autoref{fig:decision-tree}).
We show that the user mental model adaptations based on truthful understanding or untruthful rationalization both generate knowledge~\cite{pirolli2005sensemaking} and thus need to be carefully audited.\\

\noindent\textbf{Towards Truthful Understanding --} The primary goal of the explanation process is for the user to reach a \textbf{truthful understanding} of the model behavior (see~\autoref{fig:decision-tree}--1). 
In an ideal explainability scenario, a model explanation would have a \textit{high fidelity} (see~\autoref{fig:decision-tree}--X), i.e., it would match the language model behavior. Moreover, the explanation would be \textit{complete}, covering all the different model's properties. Even if we would have this ideal situation (which is obviously not a simple task to achieve), there would be another important factor that influences the success of the explainability process, i.e., the user's mental model.
If the explanation matches the user's mental model (e.g., the explanation confirms that what they know about language models and/or language), then they most likely would keep their mental model (see~\autoref{fig:decision-tree}--A) and gain some understanding of the  model's behavior.
If the explanation differs from the user's expectations, the truthful understanding could be reached only if the user is willing to adapt their mental model to match the high fidelity explanation (see~\autoref{fig:decision-tree}--B). 
The explanation should thus use an appropriate language that is legible for the target user, and it should clearly illustrate the differences between the way the language model captures natural language properties and the particular target user with their specific knowledge and expertise.
This case also covers users who are learning about language models for the first time and with no prior conception (and thus no expectation of a mental model match) willing to accept the explanation, fostering a better understanding of the language model behavior.

\begin{figure}[b!]
    \vspace{-1em}
    \centering
    \includegraphics[width=\linewidth]{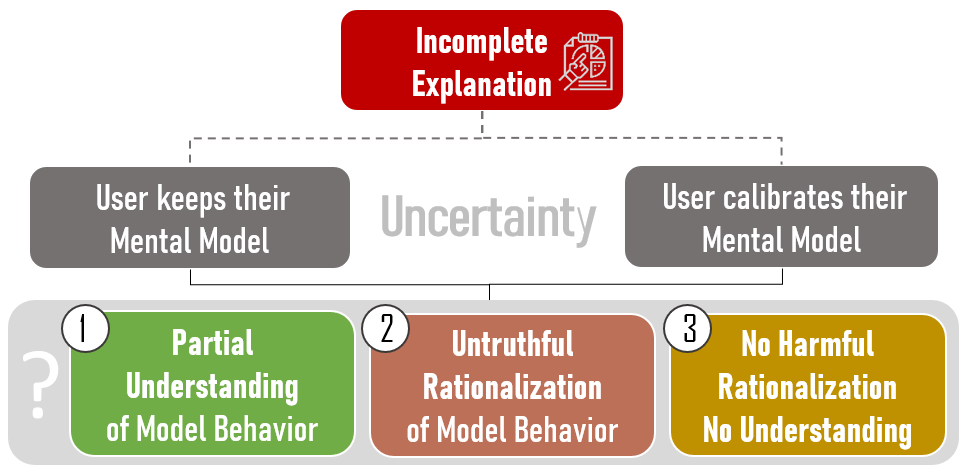}
    \caption{In reality, many NLP explanations are incomplete, i.e., they partially explain one phenomenon in isolation. Since partial explanations leave empty space for rationalization, there is a high uncertainty in the outcome of the explainability process. }
    \label{fig:completeness}
\end{figure}

\noindent\textbf{People Tend to Rationalize --} 
If the high fidelity explanation does not match the user's expectations, and they are not willing to adapt their mental model (see~\autoref{fig:decision-tree}--C), there is another possible outcome -- the user begins to \textit{rationalize} (see~\autoref{fig:decision-tree}--2). 
One of the reasons might be a lack of trust in the underlying model or the explanation process as such. 
The mismatch between the explanation outcomes and one's expectations may raise doubts about the model's learning capabilities, as well as the explanation correctness/effectiveness.
The reason for users tending to rationalize when interpreting language model outcomes lies in the nature of the input data, i.e., the natural language.
In particular, natural language, i.e., text has a meaning (also known as \textit{semantics}). 
Semantics in some cases are difficult to represent through computational methods.
For instance, \cite{sheth2005semantics} define three types of semantics, i.e., implicit, formal, and powerful semantics, and write that not all of them are machine processable.
For instance, the powerful semantics allow uncertainty permitting not only inductive but also abductive reasoning, i.e., making conclusions from what one knows rather than what one observes.
Such semantics are difficult or even impossible to learn and represent by a language model.
Thus, there can be a potential mismatch between what the user knows about the language and what they observe in the language model explanations.
The user can also misinterpret explanations due to their abductive reasoning.
The rationalization can  also occur in situations, in which the XAI provides \textit{low fidelity explanations} (see~\autoref{fig:decision-tree}--Y).
If the user is not skeptical about such explanations and decides to falsely calibrate their mental model (see~\autoref{fig:decision-tree}--D), the outcome is dangerous.
Previous studies have shown that explanations of language models, even when wrong, can increase human trust in machine predictions; e.g., in experiments showing low-accuracy statements or random heatmaps~\cite{10.1145/3287560.3287590}.

A less severe outcome in the low fidelity case occurs when the users are \textit{skeptical} and don't change their mental model. Here, their vigilance will keep users from accepting a harmful rationalization, but no model behavior understanding will be achieved  (see~\autoref{fig:decision-tree}--3). 
Although this outcome does not directly cause any harm,  it hinders users from understanding the true potential of the models and might influence their trust in future explanations.

Lastly, sometimes the explanation fidelity is \textit{unknown}; in such cases the effect of the explanation  is \textit{uncertain}. This is  similar to the case of \textit{incomplete explanations}, as  described in detail below.\\

\noindent\textbf{Incomplete Explanations Cause Uncertainty --}
Commonly, model explanations are tailored toward explaining one or a few phenomena in the black-box model at a time.
At the same time, explanations are commonly designed to show a result of one (e.g., attribution method) or a few methods in isolation.
As depicted in~\autoref{fig:completeness}, such incomplete explanations introduce uncertainty.
The outcome of incomplete explanations can be threefold. Although two outcomes are less harmful, i.e., either these explanations generate partial understanding of the model's behavior (see~\autoref{fig:completeness}--1) or no harm at all (see~\autoref{fig:completeness}--3), there is also a possibility that the \textit{missing information} will prompt users to untruthful rationalization (see~\autoref{fig:completeness}--2). 
Since the generation of complete explanations is difficult (if not impossible in some cases), most of the current explanation processes contain a high uncertainty in their outcome.
Thus, we emphasize the need for a more human-centered explanation process where the explanations clearly highlight the differences between the users' understanding of language and the language model behavior and where they are communicated in a way that allows the users to keep or calibrate their mental models with high certainty.

\section{Supporting Mental Model Calibration}
In the following, we describe actionable steps that can be undertaken to support users in gaining truthful understanding of language model behavior.\vspace{.5em}

\noindent\textbf{Avoiding False Rationalization --} 
In order to make the explanation process successful, we need to avoid several factors: (1) low fidelity explanations, (2) incomplete explanations, and (3) detachment from users' mental models. To create high fidelity explanations, we need to create means and metrics to appropriately measure their quality~\cite{guidotti2018survey}. The design of the explanations must be motivated by language model inner-workings; the user should get the chance to get insights into relevant model properties. The explanations must use language that is familiar to the users. We need to first understand the user's mental models and expectations about a phenomenon before we try to explain it, ideally relying on metaphors familiar to the user~\cite{jentner2018minions}.\vspace{.5em} 

\noindent\textbf{Catch and Counteract Pitfall Scenarios --} To ensure that users don't fall into a rationalization deadlock, we need to implement means to detect when the explanations have brought users to the wrong \textit{path}. Verification steps of that what users understand from the explanations could be integrated in the explainability process. In particular, from time to time, we could test what the user has understood about the model's working mechanisms, e.g., using verification questions~\cite{strategies2019}. If the generated insights are wrong, we would need to correct and educate the user, possibly adapting the used explanation method.\vspace{.5em} 

\noindent\textbf{Explanation as an Adaptive Process --} Explaining a phenomenon is as much about choosing the right methods as it is about enabling the person receiving the explanation to  gradually adapt their mental model. Building XAI pipelines that consist of smaller explainer units has been suggested by \cite{spinner2020explainer} to allow for the 
progressive development of explanations based on the user's understanding and tasks. 
Further, we can enhance this progressive process by deploying co-adaptive guidance and analysis strategies  \cite{sperrle2021co} to capture and react to varying users' mental models -- if the explanation system detects that the user misunderstands the presented explanations, the system could present a different type of explanation or enhance the explanation with more details.  \vspace{.5em}

\noindent\textbf{Contrastive Explanations -- } 
Top-down explanations is a common practice in XAI research. Researchers define hypotheses and test, e.g., through probing classifiers~\cite{belinkov2018,conneau2018-probing}, whether the model captures  specific phenomena. Probing classifiers are typically black-box models and, thus, such explanations can lead to harmful rationalizations. Contrasting such explanations with bottom-up  or counterfactual reasoning  can lead to more nuanced perspectives, e.g., by unfolding new specificities about a model's behavior.
These explanations are especially important when users have strong mental models about related phenomena, such as on language model learning mechanisms, requiring them to undergo a necessary calibration of  their mental models.
Hence, we might want to consider  designing systems that support open-ended exploration, as well as diverse contrastive and bottom-up explanations.\vspace{.5em}

\noindent\textbf{Human-like Explanations -- } \cite{10.1145/3278721.3278736} has introduced a technique called \textit{AI Rationalization} for explanation generation that translates internal state-actions of an autonomous agent into natural language. They argue that such a human-like communication is more accessible and intuitive to users. 
We believe that such explanations are particularly relevant for user groups whose mental models about a specific phenomenon differs from the way language models depict it.
In these cases, pure statistical explanations, e.g., saliency scores, are incomplete and insufficient to provide truthful understanding. 
Thus, we might consider extending statistical explanations with explanations in natural language that translates which features are considered as important for prediction making and why in a simple and accessible way~\cite{sevastjanova-verbal}.\vspace{.5em}

\noindent\textbf{Collaborative Learning --} 
To ease the learning process, we could benefit from collaborative environments where multiple users work together. 
Collaborative learning  motivates people to share their mental models of language and, thus, enable them to learn from both, the system and their peers  in the explanation environment.\vspace{.5em}

\section{Conclusion}
This paper discusses the need for proper human-centered design solutions for NLP explainability. 
Our thought experiment  illustrated the  explainability process and its possible outcomes under the consideration of three factors; explanation completeness, explanation quality, and user mental models. 
We argue that only complete, high fidelity explanations that match or calibrate user mental models will create user truthful understanding in language model behavior. 
To support users to gain truthful understanding of
language model behavior, we introduce several human-centered solutions.

In the future, we will work on quantifying the effects of rationalizations through different user studies using explainability interfaces. 
 Additionally, we aim to study the differences of explaining intelligible versus non-intelligible data, specifically the effects of intelligibility on rationalization. 
 
\bibliographystyle{named}
\bibliography{ijcai22}
\end{document}